\documentclass[referee,pdflatex,sn-mathphys-num]{sn-jnl}

\usepackage{graphicx}%
\usepackage{multirow}%
\usepackage{amsmath,amssymb,amsfonts}%
\usepackage{amsthm}%
\usepackage{mathrsfs}%
\usepackage[title]{appendix}%
\usepackage{xcolor}%
\usepackage{textcomp}%
\usepackage{manyfoot}%
\usepackage{booktabs}%
\usepackage{algorithm}%
\usepackage{algorithmicx}%
\usepackage{algpseudocode}%
\usepackage{listings}%
\usepackage{makecell}
\usepackage{parskip}
\usepackage{tikz}
\usepackage{fdsymbol}
\usepackage{pifont}

\theoremstyle{thmstyleone}%
\theoremstyle{thmstyletwo}%

\theoremstyle{thmstylethree}%

\begin{document}

\title[drGT: Attention-Guided Gene Assessment of Drug Response Utilizing a Drug-Cell-Gene Heterogeneous Network]{drGT: Attention-Guided Gene Assessment of Drug Response Utilizing a Drug-Cell-Gene Heterogeneous Network}


\author[1,2,3]{\fnm{Yoshitaka} \sur{Inoue}}\email{inoue019@umn.edu}
\author[1]{\fnm{Hunmin} \sur{Lee}}\email{lee03915@umn.edu}
\author[4]{\fnm{Tianfan} \sur{Fu}}\email{futianfan@gmail.com}
\author[1]{\fnm{Rui} \sur{Kuang}}\email{kuang@umn.edu}
\author*[2,3]{\fnm{Augustin} \sur{Luna}}\email{cannin@gmail.com}

\affil[1]{\orgdiv{Department of Computer Science and Engineering}, \orgname{University of Minnesota}, \orgaddress{\street{200 Union Street SE}, \postcode{55455}, \state{MN}, \country{US}}}

\affil[2]{\orgdiv{Computational Biology Branch}, \orgname{National Library of Medicine}, \orgaddress{\street{8600 Rockville Pike}, \postcode{ 20894}, \state{MD}, \country{US}}}

\affil[3]{\orgdiv{Developmental Therapeutics Branch}, \orgname{National Cancer Institute}, \orgaddress{\street{9000 Rockville Pike}, \postcode{20892}, \state{MD}, \country{US}}}

\affil[4]{\orgdiv{School of Computer Science}, \orgname{Nanjing University}, \orgaddress{\street{22 Hankou Road}, \city{Nanjing}, \postcode{210093}, \state{Jiangsu}, \country{China}}}


\abstract{ 

\textbf{Background:} For translational impact, both accurate drug response prediction and biological plausibility of predictive features are needed. We present \textit{drGT}, a heterogeneous graph deep learning model over drugs, genes, and cell lines that couples prediction with mechanism-oriented interpretability via attention coefficients (ACs). 

\textbf{Results:} We assess both \emph{predictive generalization} (random, unseen-drug, unseen-cell, and zero-shot splits) and \emph{biological plausibility} (use of text-mined PubMed gene-drug co-mentions and comparison to a structure-based DTI predictor) on GDSC, NCI60, and CTRP datasets. Across benchmarks, \textit{drGT} consistently delivers \textbf{top regression performance} while maintaining \textbf{competitive classification accuracy} for drug sensitivity. Under random 5-fold cross-validation, \textit{drGT} attains an AUROC of up to 0.945 (3rd overall) and an $R^2$ up to 0.690, outperforming all baselines on regression. In leave-one-out tests for unseen cell lines and drugs, \textit{drGT} achieves AUROCs of 0.706 and 0.844, and $R^2$ values of 0.692 and 0.022, the only model yielding positive $R^2$ for unseen drugs. In zero-shot prediction, \textit{drGT} achieves an AUROC of 0.786 and a regression $R^2$ of 0.334, both representing the highest scores among all models. For interpretability, AC-derived drug-gene links recover known biology: among 976 drugs with known DTIs, 36.9\% of predicted links match established DTIs, and 63.7\% are supported by either PubMed abstracts or a structure-based predictive model. Enrichment analyses of AC-prioritized genes reveal drug-perturbed biological processes, providing pathway-level explanations. 

\textbf{Conclusions:} \textit{drGT} advances \emph{predictive generalization} and \emph{mechanism-centered interpretability}, offering \textbf{state-of-the-art regression accuracy} and literature-supported biological hypotheses that demonstrate the use of graph learning from heterogeneous input data for biological discovery. Code: \url{https://github.com/sciluna/drGT}. 
}



\maketitle

\section{Background}\label{sec1}

Drug sensitivity arises from complex interactions among cellular states, disease-induced alterations, and molecular components~\citep{chang2019integrated}. Identifying biomarkers that capture these relationships is crucial for linking drug response to underlying biological mechanisms~\citep{chen2021data}. Deep learning (DL) algorithms have shown strong capability to model such non-linear dependencies and predict biological responses from large-scale molecular data~\citep{Mahmud2020Deep}.

Despite advances in applying DL to drug discovery~\cite{lu2022cot}, a key challenge that remains is the “black box” nature of these methods. Although DL models achieve high predictive accuracy, their internal decision-making often lacks transparency, raising concerns about their reliability and limiting their utility for scientific discovery~\cite{corso2023holistic,michael2018visible}. In biomedical contexts, performance metrics alone are insufficient; models must also provide biologically plausible explanations to be useful in hypothesis generation~\citep{chen2022minimum}.

These limitations have motivated the development of interpretable DL architectures~\cite{kengkanna2023enhancing}, particularly in drug discovery~\cite{fu2022hint, yang2022active}. Among the most promising approaches is the attention mechanism, introduced in the Transformer architecture~\cite{vaswani2017attention}, which employs trainable attention coefficients (ACs) to quantify the importance of individual components, thereby improving interpretability. While words in a sentence convey a coherent idea, biological regulation depends on complex interconnections among molecular entities, making graph-based representations more suitable for modeling biological systems. Graph Neural Networks (GNNs) have been proposed to capture such relational dependencies, and Graph Transformers (GTs) extend this paradigm by applying global attention to graph data~\cite{shi2020masked}. Recent work has explored graph attention and meta-learning for low-data scenarios, such as Meta-GAT architecture~\cite{lv2023meta} and the Meta-MolNet benchmark~\cite{lv2023meta}, both of which focus on few-shot molecular property prediction. 

\begin{figure*}[!ht]
\centering\includegraphics[width=\textwidth,keepaspectratio]{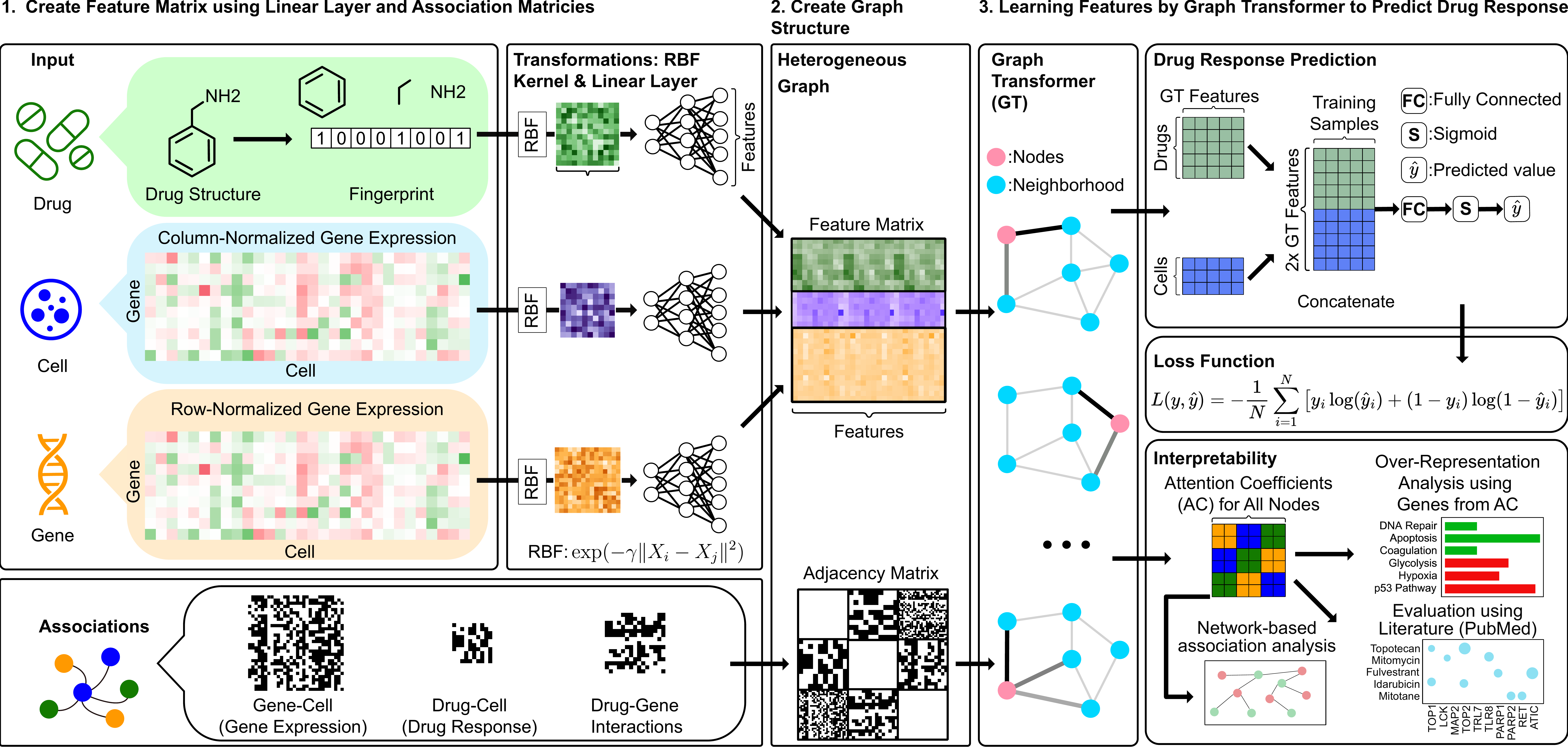}
\caption{drGT overview. A heterogeneous graph is built using drug structures and gene expression data. Each input type is transformed using an RBF kernel, followed by linear layers to project features into a common space. These transformed inputs are then combined into a unified feature matrix, which serves as input to the GT layers along with an adjacency matrix. For prediction, the GT layer output is concatenated to align with the drug response data. Following fully connected (FC) layers, a sigmoid function generates predicted values $\hat{y}$, which are then inputs for the binary cross-entropy (BCE) loss. To assess interpretability, ACs are utilized in conjunction with over-representation analysis and network-based association analysis.}
\label{fig:1}
\end{figure*}

We propose an interpretable DL framework, drGT, which leverages GTs to process a large-scale heterogeneous graph integrating drug compounds, cell lines, and genes. By applying global attention over this structured network, the GT layers learn embeddings that capture complex drug-gene-cell dependencies. Unlike previous approaches that rely on homogeneous graphs or lack mechanism-oriented interpretability, drGT provides fine-grained attribution through attention coefficients (ACs), enabling direct assessment of gene contributions to drug response.

We apply drGT to two primary tasks: (i) predicting drug response and (ii) interpreting gene importance. We evaluate the model in both binary classification and regression settings across three complementary scenarios: (1) random masking for matrix reconstruction, (2) leave-one-out prediction of unseen drugs or cell lines, and (3) zero-shot transfer across datasets. Across all evaluations, drGT achieves state-of-the-art (SOTA) or competitive performance while maintaining biologically meaningful interpretability through ACs that highlight the relative importance of specific nodes and drug-gene associations.

We analyze ACs assigned to gene nodes to explain drug responses. Unlike existing interpretable methods such as SubCDR~\cite{liu2023subcomponent} and DRPreter~\cite{shin2022drpreter}, drGT offers finer-grained, edge-level interpretability by leveraging ACs on drug-gene connections. This allows quantifying each gene’s contribution to a drug’s response, yielding biologically meaningful and actionable insights. To validate the biological plausibility of these associations, we benchmark AC-derived drug-gene links against PubMed co-occurrence data and predictions from DeepPurpose~\cite{huang2020deeppurpose}, a state-of-the-art DTI model, confirming both biological and computational consistency. Moreover, relevant biological processes are described using the ACs as part of an over-representation analysis.

\section{Methods}\label{sec11}

drGT proceeds in four main steps: (1) construction of a heterogeneous drug-cell-gene graph from drug structures, gene expression, drug response, and drug-target interaction data; (2) transformation of drug, cell line, and gene features into a unified latent space using kernel-based similarity and linear projections; (3) learning of node representations and drug-cell response prediction using Graph Transformer layers; and (4) extraction of attention coefficients to interpret feature (gene) contributions and assess biological plausibility via literature and external models. (see Fig.~\ref{fig:1})

\subsection{Building Input Matrix}

We collected gene expression data from NCI60~\cite{shoemaker2006nci60}, GDSC1/2~\cite{yang2012genomics}, and CTRP~\citep{rees2016correlating} via rcellminer~\citep{luna2016rcellminer}. Drug response half-maximal inhibitory concentration (IC50) values were obtained from PharmacoDB~\citep{smirnov2018pharmacodb}, and drug-target interaction (DTI) data from DrugBank~\citep{wishart2018drugbank}.

The drug-response matrix is defined as \(X_{DR} \in \mathbb{R}^{n \times m}\), where \(n\) is the number of drugs and \(m\) the number of cell lines. Gene expression is represented as \(X_{GE} \in \mathbb{R}^{m \times l}\), with \(l\) genes. The DTI matrix \(X_{DTI} \in \mathbb{R}^{n \times l}\) is zero-padded for unknown interactions. For molecular features, SMILES strings were converted to 2048-bit Morgan fingerprints~\citep{morgan1965generation} using RDKit~\citep{landrum2016rdkit}. These vectors form the matrix \(X_{MF} \in \mathbb{R}^{n \times o}\), where \(o\) denotes the fingerprint dimension.

\subsubsection{Preprocessing}

For classification, IC50 values were log10-transformed and filtered to the central 95\% range (2.5-97.5 percentiles). Motivated by prior studies that adopt drug-wise definitions of sensitivity and resistance~\citep{iorio2016landscape}, we applied per-drug Z-score normalization and binarized responses using a margin-based threshold:
{\setlength{\abovedisplayskip}{5pt}
\setlength{\belowdisplayskip}{5pt}
\begin{equation*}
\text{Label} =
\left\{
\begin{array}{ll}
\text{Sensitive} & \text{if } x < \mu - \sigma \\
\text{Resistant} & \text{if } x > \mu + \sigma  \\
\text{Uncertain} 
\end{array}
\right.
\end{equation*}
}
where \(\mu\) and \(\sigma\) denote the mean and standard deviation of IC50 values for each drug. Pairs labeled ``Sensitive'' or ``Resistant'' are included as binary labels (i.e., ``Sensitive'' is 1 and ``Resistant'' is 0.)

For the regression task, IC50 values were filtered to remove invalid entries (0 or $\infty$), transformed into pIC50 ($-{\log_{10}}(\text{IC50})$), and clipped to 0-15 to exclude outliers. Drug-cell pairs with fewer than ten valid measurements were discarded. Preprocessed dataset stats are described in Table~\ref{stats_cls_reg}.

\begin{table}
\centering
\caption{Summary of dataset characteristics used in classification and regression settings. The table reports the number of drugs, cell lines, drug-response pairs, known drug-target interactions, and selected genes for each dataset.}
\label{stats_cls_reg}
\begin{tabular}{l|rrrr}
\toprule
Classification & NCI60 & GDSC1 & GDSC2 & CTRP \\ \midrule
Drugs & 976 & 300 & 153 & 470 \\
Cell Lines & 59 & 925 & 654 & 807 \\
Total Drug-Response & 14,312 & 31,084 & 10,113 & 46,313 \\
Known Drug-Target & 572 & 722 & 313 & 783 \\
Genes & 2,489 & 2,106 & 2,040 & 2,160 \\
\midrule
Regression & NCI60 & GDSC1 & GDSC2 & CTRP \\
\midrule
Drugs & 1,023 & 193 & 95 & 190 \\
Cell Lines & 11 & 943 & 601 & 793 \\
Total Drug-Response & 10,723 & 43,805 & 13,416 & 34,350 \\
Known Drug-Target & 126 & 643 & 266 & 588 \\
Genes & 2,376 & 2,082 & 2,022 & 2,111 \\
\bottomrule
\end{tabular}
\end{table}

\subsubsection{Feature Construction}

To construct the feature matrices, we generated pairwise similarity matrices from \(X_{GE}\) and \(X_{MF}\). Specifically, we computed cell line (\(S_c\)), gene (\(S_g\)), and drug (\(S_d\)) similarities using an RBF kernel defined as
\[
S_{ij} = \exp\!\left(-\gamma \lVert X_i - X_j \rVert^2\right),
\]
where \(\gamma = 1/\text{length}(X_i)\). The kernel was applied to \(X_{GE}\), \(X_{GE}^\top\), and \(X_{MF}\) to obtain \(S_c\), \(S_g\), and \(S_d\), respectively.

\subsubsection{Graph Construction}

We constructed adjacency matrices from \(X_{DR}\), \(X_{GE}\), and \(X_{DTI}\). The drug-cell association matrix \(A_{dc}\) is obtained from the original drug-response matrix \(X_{DR} \in \mathbb{R}^{n \times m}\). For classification tasks, binary labels are used, whereas for regression, the continuous response values are directly assigned as edge attributes. To prevent information leakage, all test-set entries  \(X_{DR}\) are masked before constructing \(A_{dc}\); specifically, the corresponding values are replaced with zeros so that no test information is included in the unified graph used by the Graph Transformer. We note that only drug-cell response edges (\(A_{dc}\)) are masked for evaluation. The cell-gene (\(A_{cg}\)) and drug-gene (\(A_{dg}\)) matrices are retained for both training and test entities. This corresponds to a transductive learning setting, in which biological information (e.g., gene expression and curated drug-target annotations) is assumed to be available prior to response screening. Our objective is to predict drug-cell response values within a known biological context, rather than to infer drug-gene or cell-gene relationships.

The cell-gene association matrix \(A_{cg}\) is derived from the gene expression matrix \(X_{GE} \in \mathbb{R}^{m \times l}\). We first standardize gene expression values across cell lines and retain only positive values, setting all non-positive standardized entries to zero. To assess the robustness of this thresholding strategy, we also evaluated variants that preserved both the highly and lowly expressed genes (e.g., the top and bottom 5\% or 25\% per cell line). The results were consistent across these alternatives, supporting the use of the high-expression-based formulation.

The drug-gene association matrix \(A_{dg} \in \mathbb{R}^{n \times l}\) is constructed from the drug-target interaction data \(X_{DTI}\), where rows correspond to drugs and columns correspond to genes. Known drug-gene interactions are assigned a value of 1, whereas unobserved interactions are set to 0.

Using the three association matrices \(A_{dc}\), \(A_{cg}\), and \(A_{dg}\), 
we construct the unified adjacency matrix \(A\) as:
\begin{equation} \label{eq:final_matrix}
    A = 
    \begin{bmatrix}
        \mathbf{0} & A_{dc} & A_{dg} \\
        A_{dc}^\intercal & \mathbf{0} & A_{cg} \\
        A_{dg}^\intercal & A_{cg}^\intercal & \mathbf{0}
    \end{bmatrix},
\end{equation}
where \(A \in \mathbb{R}^{(n+m+l) \times (n+m+l)}\) represents the unified adjacency matrix over drugs, cells, and genes. Here, \(\mathbf{0}\) denotes zero matrices of appropriate dimension.

\subsection{drGT Model}

\paragraph{Graph Transformer Layer.}
To specify the operation used inside each GNN block, we briefly define the Graph Transformer (GT) layer. We adopt the \texttt{TransformerConv} operator from PyTorch Geometric~\citep{fey2019fast}, which extends Transformer-style attention to graph-structured data. A GT layer performs masked multi-head self-attention (MHA) over the edges of the unified adjacency matrix $A$, incorporates edge features into the attention mechanism, and applies a feed-forward transformation with normalization, activation, and optional residual connections. Below, we detail how attention is computed inside the GT layer.

\paragraph{Attention Computation.}
Given node features $H \in \mathbb{R}^{N \times d}$, the GT layer computes query, key, and value matrices as
\[
Q = H W_Q,\qquad 
K = H W_K,\qquad
V = H W_V,
\]
where $W_Q$, $W_K$, and $W_V$ are learnable parameters.  Attention is computed only over edges $(i,j)$ present in the adjacency matrix $A$. Each edge carries an edge feature $e_{ij}$ (e.g., drug-cell response, drug-gene interaction, or gene expression after thresholding), which is incorporated into the key and value projections:
\[
K_j \leftarrow K_j + W_E e_{ij}, \qquad
V_j \leftarrow V'_E e_{ij}.
\]
The unnormalized attention scores and the softmax-normalized attention coefficients are defined as
\[
e_{ij} = \frac{Q_i K_j^\top}{\sqrt{d_k}}, \qquad
\alpha_{ij} = 
\frac{\exp(e_{ij})}{\sum_{k \in \mathcal{N}(i)} \exp(e_{ik})},
\]
where $\mathcal{N}(i)$ denotes the neighbors of node $i$ in $A$.  

For $H$ attention heads, the aggregated output is
\[
\mathrm{MHA}(H)_i
=
\big\Vert_{h=1}^H 
\sum_{j \in \mathcal{N}(i)}
\alpha^{(h)}_{ij} V^{(h)}_j,
\]
followed by a linear transformation and an optional residual connection. This transformed representation is then passed through a GraphNorm, ReLU, and Dropout.

The resulting GT layer is applied to the inputs of our heterogeneous graph as follows:

\begin{equation}\label{eq:z1}
\begin{aligned}
Z_1
&= \mathrm{Dropout}\Big(
      \mathrm{ReLU}\Big(
      \mathrm{GraphNorm}\Big(
        \mathrm{GT}\Big(
          \begin{bmatrix}
              \mathrm{Linear}_d(S_d) \\
              \mathrm{Linear}_c(S_c) \\
              \mathrm{Linear}_g(S_g)
          \end{bmatrix},
          A
        \Big)
      \Big)
      \Big)
    \Big).
\end{aligned}
\end{equation}

where $\mathrm{Dropout}$ denotes the dropout layer, $\mathrm{ReLU}$ the rectified linear activation, $\mathrm{GraphNorm}$ the graph normalization layer~\citep{cai2021graphnorm}, $\mathrm{GT}$ the Graph Transformer layer, and $\mathrm{Linear}_d$, $\mathrm{Linear}_c$, and $\mathrm{Linear}_g$ are individual linear layers that map drug, cell-line, and gene features into the same dimensionality. This normalization reduces scale disparities in node representations, particularly under heterogeneous degree distributions, and helps prevent dominance by high-degree nodes during attention aggregation, contributing to more stable training~\citep{cai2021graphnorm}.

Subsequent GNN layers repeat the same operation as equation~\eqref{eq:z1}, producing intermediate representations $\{Z_\ell\}_{\ell=1}^{L}$, where $L$ (the number of GNN layers) is a hyperparameter. The final representation $Z_L$ is used for prediction. Drug-cell line associations are obtained by concatenating their embeddings from $Z_L$ and passing the result through fully connected (FC) layers followed by a sigmoid activation: \[ \hat{y} = \mathrm{sigmoid}\!\left( \mathrm{FC}\!\left( Z_{L,d} \,\mathbin{\Vert}\, Z_{L,c} \right) \right), \] where $Z_{L,d}$ and $Z_{L,c}$ denote the drug and cell-line embeddings. For classification, we use binary cross-entropy (BCE): \[ L_{\mathrm{BCE}}(y, \hat{y}) = -\frac{1}{N}\sum_{i=1}^{N} \left[ y_i \log \hat{y}_i + (1-y_i)\log(1-\hat{y}_i) \right]. \] For regression, the model is trained with mean squared error (MSE): \[ L_{\mathrm{MSE}}(y, \hat{y}) = \frac{1}{N}\sum_{i=1}^{N}(y_i - \hat{y}_i)^2. \] The network is trained using Adam~\citep{kingma2014adam} and implemented with PyTorch 0.13.1~\citep{paszke2019pytorch} and PyTorch Geometric 2.2.0~\citep{fey2019fast}. Hyperparameters are tuned with Optuna~\citep{akiba2019optuna}, exploring hidden sizes, dropout rates, layer counts, attention heads, activation functions, optimizers, normalization schemes, learning rates, and epochs.

\subsection{Interpretability}

To provide mechanism-oriented interpretability, drGT leverages the attention coefficients (ACs) produced by the GT layers. All interpretability analyses use ACs computed on the validation split to avoid stochastic fluctuations introduced by dropout during training. For interpretation, unknown drug-gene interactions are assigned a soft edge value of $0.5$ so that these edges are retained in the attention computation, allowing attention to diffuse to potentially relevant genes while preserving the graph structure.

\paragraph{Interpretation of Drug-Gene Relationships.}
For each drug node $i$, we extract the attention coefficients on its incident drug-gene edges. Given multi-head coefficients $\alpha^{(h)}_{ij}$ from the final GT layer, we compute a unified attention score by averaging across heads: \[ \hat{\alpha}_{ij} = \frac{1}{H} \sum_{h=1}^{H} \alpha^{(h)}_{ij}. \] This score is used consistently across all interpretability analyses. Genes are ranked by $\hat{\alpha}_{ij}$, and for pathway-level analysis we select the top-$k$ genes (default $k=100$), providing sufficient coverage for enrichment.

\paragraph{Evaluation of Attention-Derived Drug-Gene Associations.}
To evaluate whether the model prioritizes biologically meaningful drug-gene relationships, we extract for each drug $i$ the top-$5$ genes ranked by the unified attention score $\hat{\alpha}_{ij}$. These top-$5$ genes are used consistently across all external validation analyses.

To assess literature support, we queried PubMed via NCBI ESearch using the pattern \texttt{"<drug\_name>" AND "<gene\_symbol>"}, and recorded a co-occurrence when at least one abstract was returned. Drug names were standardized using NSC identifiers, and gene symbols followed the HGNC convention. Known drug-target interactions were obtained from DrugBank.

\paragraph{External Validation of AC-Derived Drug-Gene Associations by Machine Learning.}
To evaluate whether AC-derived drug-gene associations align with predictions from an independent machine learning model, we applied DeepPurpose~\cite{huang2020deeppurpose}, a state-of-the-art DTI predictor pre-trained on BindingDB~\cite{liu2007bindingdb}. For each drug, the top-5 AC-ranked genes were compared against DeepPurpose scores computed using the default parameters for DeepPurpose, providing an orthogonal computational assessment of the plausibility of the predicted associations.

\paragraph{Higher-Level Interpretability.}

To reveal broader biological structure, we performed pathway enrichment using the top-$k$ AC-ranked genes (default $k=100$) derived from $\hat{\alpha}_{ij}$. We conducted an over-representation analysis (ORA) using the ACs attributed to the genes associated with the drugs. For each drug, the top 100 genes with the highest ACs were analyzed using the gseapy package~\citep{fang2023gseapy} and the MSigDB Hallmark 2020~\citep{liberzon2015molecular} collection of 50 gene sets accessed via gseapy. This enables drGT to map attention-derived gene signatures to transcriptional programs and molecular processes associated with drug sensitivity or resistance.

\subsection{Comparison Models and Baseline Architectures}
To contextualize the performance of drGT, we compared it against six widely used SOTA drug response prediction (DRP) models: DeepDSC~\cite{li2019deepdsc}, MOFGCN~\cite{peng2021predicting}, GraphCDR~\cite{liu2022graphcdr}, SubCDR~\cite{liu2023subcomponent}, NIHGCN~\cite{peng2022predicting}, and DRPreter~\cite{shin2022drpreter} (Table~\ref{tab:baseline}). These methods span a broad design space, including feed-forward networks (FFNs), GNNs, biologically grounded pathway models, and interaction-based frameworks. All baseline implementations follow their originally reported hyperparameter configurations. For methods that incorporate additional omics modalities (e.g., mutation, methylation, or copy number variation), we retained the same input configurations as specified in their original studies.

Each baseline makes distinct architectural assumptions that lead to fundamental differences in their representational capacity:

DeepDSC employs FFNs and an autoencoder to encode fingerprints and gene-expression vectors. Its architecture is vector-based and lacks any mechanism to capture relational structure between entities. As a result, it cannot model drug-cell interactions beyond the concatenation of independent representations.

GraphCDR incorporates graph convolutional networks (GCNs)~\citep{kipf2016semi} for molecular graph encoding and dense networks for cell-line omics, and then both cell line and drug embeddings are projected into a drug-cell bipartite graph for reasoning. However, GraphCDR processes drugs and cells using separate encoders and does not include gene nodes or biological associations, limiting cross-modality information sharing. Furthermore, because the bipartite edges encode only drug-cell response labels, the model cannot exploit gene-level structure or drug-gene relationships.

MOFGCN applies unified GCNs on a similarity-association graph whose nodes include drugs and cell lines. Although MOFGCN integrates multi-omics data through similarity fusion, its graph contains no gene nodes, no drug-gene edges, and no biological pathways. Consequently, the model can propagate information only across phenotype-level similarities rather than mechanistic biological interactions.

NIHGCN extends bipartite GNNs by adding neighborhood-interaction (NI) layers that capture element-wise interactions between drug and cell embeddings. While this improves the modeling of local dependencies within the drug-cell bipartite graph, NIHGCN still lacks gene-level node structure and does not incorporate pathway-level information.

DRPreter and SubCDR explicitly introduce biologically meaningful structures to enhance interpretability. DRPreter constructs pathway-level subgraphs from PPI networks and combines them with drug molecular graphs via a transformer, enabling pathway-level attribution. SubCDR decomposes drug molecules into Breaking of Retrosynthetically Interesting Chemical Substructures (BRICS)-derived substructures and cell lines into curated gene subsets, and models their pairwise interactions in a subcomponent interaction graph. Although these biologically motivated designs improve interpretability, both models remain limited to drug-cell representations and do not construct an integrated drug-cell-gene graph, preventing joint reasoning across all three entity types.

In contrast to these approaches, drGT represents drugs, cell lines, and genes within a single unified heterogeneous graph. A Graph Transformer encoder with typed edges (e.g., drug-gene, cell-gene, and gene-gene relations) enables drGT to simultaneously capture: (i) molecular graph structure of drugs, (ii) multi-level gene relationships (e.g., pathways, co-expression, functional associations), and (iii) drug-gene and cell-gene dependencies. This unified graph representation allows drGT to model biologically grounded relational structure and cross-modal interactions in a manner that previous DRP models cannot.

Importantly, these baseline models span different prediction tasks (binary sensitivity classification versus continuous IC50 regression). For methods without official implementations of both settings, specifically, DeepDSC and DRPreter (classification) and GraphCDR and MOFGCN (regression), we follow their original architectures and switch the output loss to either binary cross-entropy (BCE) or mean squared error (MSE), ensuring a fair and consistent evaluation protocol across tasks.

\begin{sidewaystable}
\begingroup
\setlength{\parskip}{0pt} 
\small
\setlength{\tabcolsep}{5pt}
\centering
\caption{Comparison of representative drug response prediction models. ``Interp.'' denotes model interpretability, defined as the ability to reveal biologically meaningful factors contributing to drug response, while ``Enc.'' refers to the encoder architectures used by each model.}
\label{tab:baseline}
\begin{tabular}{l | c | c | c c | c c | p{2.8cm} | c}
\hline
\textbf{Methods} & \textbf{Year} & \textbf{Interp.} & \textbf{Drug Enc.} & \textbf{Cell Enc.} & \textbf{Drug Data} & \textbf{Cell Data} & \textbf{Graph Type} & \textbf{Task} \\
\hline
DeepDSC   & 2021 & \textcolor{red}{\ding{55}} & FFN & AutoEncoder & Fingerprint & Exp & - & Reg \\
GraphCDR  & 2022 & \textcolor{red}{\ding{55}} & GCN & Linear & Graph (SMILES) & Exp/Mut/Met & Drug-Cell bipartite graph & Binary \\
MOFGCN    & 2022 & \textcolor{red}{\ding{55}} & \multicolumn{2}{c|}{GCN} & Fingerprint & Exp/Mut/CNV & Similarity + association graph & Binary \\
NIHGCN    & 2022 & \textcolor{red}{\ding{55}} & \multicolumn{2}{c|}{GCN+NI} & Fingerprint & Exp & Drug-Cell bipartite graph & Reg/Binary \\
DRPreter  & 2022 & \textcolor[rgb]{0.13,0.55,0.13}{\ding{51}} & GIN & GAT & Graph (SMILES) & Exp & Pathway graph + Drug graph & Reg \\
SubCDR    & 2023 & \textcolor[rgb]{0.13,0.55,0.13}{\ding{51}} & GRU & CNN & Graph (BRICS) & Exp & Subcomponent interaction graph & Reg/Binary \\
drGT      & -   & \textcolor[rgb]{0.13,0.55,0.13}{\ding{51}} & \multicolumn{2}{c|}{GT} & Fingerprint & Exp & Drug-Cell-Gene unified graph & Reg/Binary \\
\hline
\end{tabular}
\endgroup
\end{sidewaystable}

\section{Results}\label{results}

\subsection{Data Summary and Processing} 

Table~\ref{stats_cls_reg} summarizes the characteristics of the four pharmacogenomic datasets used in this study. These datasets differ widely in their scale: for example, NCI60 contains a small number of cell lines (60 cell lines) but a large set of screened drugs, whereas CTRP and GDSC provide substantially larger drug-response matrices each with over 500 cell lines. The number of available drug-target interactions (DTIs) also varies, reflecting differences in experimental annotation across datasets. Such heterogeneity motivates the need for a unified modeling framework capable of accommodating diverse dataset sizes and coverage. 

For the NCI60 dataset specifically, we restricted the set of usable drugs to those belonging to a set of major CellMiner MoA categories (DNA, TUBB, Ho, Apoptosis, Acetalax, Kinase, Methylation, HSP90, HDAC, PSM, and BRD), as well as any drugs in the NCI60 that overlapped with compounds present in GDSC or CCLE. This ensures consistent drug identity across datasets and retains compounds with reliable annotation or cross-dataset overlap. Consequently, the number of NCI60 drugs used for classification and regression is smaller than the full screening panel.

The numbers reported for classification and regression differ because each task requires a distinct preprocessing pipeline. For classification, IC50 values are binarized after log-transformation and normalization, and drug-cell pairs labeled as ``uncertain'' are removed. In contrast, regression retains continuous pIC50 values, excludes invalid measurements (0 or $\infty$), and filters out drugs with fewer than 10 valid response observations. These task-specific preprocessing steps naturally lead to different sets of drugs, cell lines, and valid drug-response pairs for classification versus regression, as reflected in Table~\ref{stats_cls_reg}. 

For each dataset, we selected genes based on their variability and biological relevance. We first retained genes whose expression variability was in the top 10\% across cell lines, ensuring that the selected genes carried a meaningful signal for model training. This variability-based selection reflects a trade-off between signal preservation and computational scalability, as highly variable genes are more likely to capture drug-response-relevant transcriptional variation while excluding low-variance features that are less likely to contribute to model discrimination~\cite{parca2019modeling}. In addition, all genes appearing in the DrugBank DTI annotations were included to preserve known pharmacological interactions. After this filtering, each dataset contained approximately 2{,}000-2{,}500 genes (Table~\ref{stats_cls_reg}). 

\subsection{Experimental Design} \label{sec:5-1} 

We evaluated our model on four benchmark pharmacogenomic datasets—GDSC1, GDSC2, CTRP, and NCI60—using AUROC for classification and $R^2$ for regression. To comprehensively assess predictive performance, robustness, and generalization, we designed three complementary evaluation settings: 

\paragraph{(Test 1) Random Mask Imputation with 5-Fold Cross-Validation.} To assess reconstruction performance, we randomly masked 20\% of the observed drug-cell response values in each fold and trained the model to impute the missing entries. Five-fold cross-validation was performed by splitting the observed response matrix into training (80\%) and held-out masked entries (20\%). Masking was applied only to observed response values, and the masked entries were used exclusively for evaluation. This setting evaluates the model’s ability to capture latent drug-cell relationships and accurately infer unknown responses. 

\paragraph{(Test 2) Leave-One-Drug/Cell-Line-Out Generalization.} In contrast to Test 1, which masks individual response entries, to evaluate the ability to generalize to previously unseen drug-cell response interactions, we masked either (i) all drug response values for a given drug (LODO) across all cell lines or (ii) cell line (LOCO) response across all drugs. During training, the model did not observe any drug-cell response interaction involving the held-out drug or cell line. Evaluation was then performed on the fully masked row or column. This setting simulates pharmacogenomic scenarios in which information prior to screening (e.g., gene expression and some curated drug-target annotations) is available, but drug-cell response measurements are missing, corresponding to a transductive evaluation framework.

\paragraph{(Test 3) Cross-Dataset Zero-Shot Transfer.} To measure cross-study generalization, we adopted a zero-shot learning protocol: the model was trained on one dataset (e.g., GDSC1) and evaluated directly on another (e.g., GDSC2) without fine-tuning. This setting tests the model’s robustness to distributional shifts, batch effects, and differences in assay conditions across datasets. No explicit cross-dataset batch-effect correction was applied between GDSC1 and GDSC2. Each dataset was normalized independently during preprocessing to ensure internal consistency while preserving inter-dataset distributional differences.

\subsection{Model Performance}

\subsubsection{Performance Imputing Randomly Masked Values (Test 1)}\label{tex:test1}

\begin{sidewaystable}
\begingroup
\setlength{\parskip}{0pt} 
\small                     
\setlength{\tabcolsep}{4pt}

\centering
\caption{%
Test 1: Performance under 5-fold CV with randomly masked associations.
Each cell shows the mean (top) and standard deviation (bottom).
``Interp.'' indicates model interpretability, defined as the ability to reveal biologically meaningful factors contributing to drug response.
Column bests: \textbf{bold} (1st), \underline{underlined} (2nd), \textdagger{} (3rd).}
\label{tab:auroc_r2_combined}
\begin{tabular*}{\textheight}{@{\extracolsep\fill}l *{4}{c} *{4}{c}}
\toprule
\multirow{2}{*}{Method}
  & \multicolumn{4}{c}{Classification (AUROC)}
  & \multicolumn{4}{c}{Regression ($R^2$)} \\
\cmidrule(lr){2-5}\cmidrule(lr){6-9}
 & NCI60 & CTRP & GDSC1 & GDSC2 & NCI60 & CTRP & GDSC1 & GDSC2 \\
\midrule
DeepDSC 
& \makecell{0.488 \\ {\scriptsize (±0.015)}} 
& \makecell{0.581 \\ {\scriptsize (±0.005)}} 
& \makecell{0.591 \\ {\scriptsize (±0.006)}} 
& \makecell{0.651 \\ {\scriptsize (±0.013)}} 
& \makecell{0.482\textsuperscript{\textdagger} \\ {\scriptsize (±0.012)}} 
& \makecell{0.472\textsuperscript{\textdagger} \\ {\scriptsize (±0.017)}} 
& \makecell{\underline{0.449} \\ {\scriptsize (±0.014)}} 
& \makecell{\underline{0.503} \\ {\scriptsize (±0.025)}} \\

GraphCDR 
& \makecell{0.924 \\ {\scriptsize (±0.006)}} 
& \makecell{0.920 \\ {\scriptsize (±0.004)}} 
& \makecell{0.809 \\ {\scriptsize (±0.050)}} 
& \makecell{0.871 \\ {\scriptsize (±0.011)}} 
& \makecell{-1.099 \\ {\scriptsize (±0.062)}} 
& \makecell{-0.010 \\ {\scriptsize (±0.002)}} 
& \makecell{-0.000 \\ {\scriptsize (±0.001)}} 
& \makecell{0.000 \\ {\scriptsize (±0.001)}} \\

SubCDR   
& \makecell{0.924 \\ {\scriptsize (±0.004)}} 
& \makecell{0.873 \\ {\scriptsize (±0.022)}} 
& \makecell{0.750 \\ {\scriptsize (±0.046)}} 
& \makecell{0.780 \\ {\scriptsize (±0.024)}} 
& \makecell{-1.184 \\ {\scriptsize (±0.250)}} 
& \makecell{-0.639 \\ {\scriptsize (±0.442)}} 
& \makecell{-0.850 \\ {\scriptsize (±0.298)}} 
& \makecell{-0.915 \\ {\scriptsize (±0.323)}} \\

DRPreter  
& \makecell{0.818 \\ {\scriptsize (±0.171)}} 
& \makecell{0.922 \\ {\scriptsize (±0.013)}} 
& \makecell{0.768 \\ {\scriptsize (±0.132)}} 
& \makecell{0.872 \\ {\scriptsize (±0.007)}} 
& \makecell{\underline{0.572} \\ {\scriptsize (±0.045)}} 
& \makecell{\underline{0.547} \\ {\scriptsize (±0.027)}} 
& \makecell{0.410\textsuperscript{\textdagger} \\ {\scriptsize (±0.222)}} 
& \makecell{0.462\textsuperscript{\textdagger} \\ {\scriptsize (±0.022)}} \\

MOFGCN 
& \makecell{\underline{0.956} \\ {\scriptsize (±0.003)}} 
& \makecell{\underline{0.951} \\ {\scriptsize (±0.003)}} 
& \makecell{\underline{0.897} \\ {\scriptsize (±0.002)}} 
& \makecell{\underline{0.908} \\ {\scriptsize (±0.007)}} 
& \makecell{-0.143 \\ {\scriptsize (±0.016)}} 
& \makecell{0.098 \\ {\scriptsize (±0.008)}} 
& \makecell{0.137 \\ {\scriptsize (±0.011)}} 
& \makecell{0.168 \\ {\scriptsize (±0.009)}} \\

NIHGCN 
& \makecell{\textbf{0.965} \\ {\scriptsize (±0.002)}} 
& \makecell{\textbf{0.956 }\\ {\scriptsize (±0.002)}} 
& \makecell{\textbf{0.914} \\ {\scriptsize (±0.003)}} 
& \makecell{\textbf{0.928} \\ {\scriptsize (±0.009)}} 
& \makecell{-0.138 \\ {\scriptsize (±0.014)}} 
& \makecell{0.101 \\ {\scriptsize (±0.008)}} 
& \makecell{0.140 \\ {\scriptsize (±0.010)}} 
& \makecell{0.171 \\ {\scriptsize (±0.008)}} \\

drGT  
& \makecell{0.945\textsuperscript{\textdagger} \\ {\scriptsize (±0.003)}} 
& \makecell{0.935\textsuperscript{\textdagger} \\ {\scriptsize (±0.003)}} 
& \makecell{0.884\textsuperscript{\textdagger} \\ {\scriptsize (±0.002)}} 
& \makecell{0.896\textsuperscript{\textdagger} \\ {\scriptsize (±0.010)}} 
& \makecell{\textbf{0.690} \\ {\scriptsize (±0.017)}} 
& \makecell{\textbf{0.580} \\ {\scriptsize (±0.011)}} 
& \makecell{\textbf{0.475} \\ {\scriptsize (±0.010)}} 
& \makecell{\textbf{0.544} \\ {\scriptsize (±0.029)}} \\
\bottomrule
\end{tabular*}

\endgroup
\end{sidewaystable}

In Test 1, to assess the model's capacity to reconstruct drug sensitivity, drug-cell line associations were split into five subsets for 5-fold CV. The results are summarized in Table~\ref{tab:auroc_r2_combined}.

\textbf{Classification:} Regarding the classification task, drGT\ ranks among the top three models across four datasets, achieving an AUROC of 94.5\% on NCI60, trailing NIHGCN. Overall, NIHGCN achieved the highest scores, followed by MOFGCN. From an architectural perspective, GNN-based models (MOFGCN, NIHGCN, GraphCDR, SubCDR, DRPreter, and drGT) outperform the feedforward baseline (DeepDSC). Among these, only MOFGCN and GraphCDR integrate multi-omics features such as mutations, methylation, and copy number variation; the others rely on gene expression. Although MOFGCN incorporates multi-omics data, it does not outperform NIHGCN, indicating the difficulty of integrating omics features. Notably, among interpretable models, drGT achieved the highest AUROC across four datasets, maintaining both interpretability and strong predictive performance.

\textbf{Regression:} In the regression task, drGT achieved the highest coefficient of determination ($R^2$) across all datasets, demonstrating superior accuracy in estimating continuous drug responses. Interestingly, despite its simple architecture based solely on gene expression and chemical 2D structures, DeepDSC exhibited competitive performance, suggesting that even shallow architectures can capture essential patterns under certain conditions. Moreover, DRPreter, which also incorporates interpretability constraints, achieved moderate $R^2$ scores, reflecting the trade-off between interpretability and predictive accuracy inherent to explainable models.

\subsubsection{Performance to Unseen Drugs or Cell Lines (Test 2)}

\begin{table}[h]
\caption{Test 2: Leave-one-out evaluation on unseen drugs and cell lines in GDSC2. Results are averaged across folds, with mean AUROC and R\textsuperscript{2} (standard deviations in parentheses). `Cell' and `Drug' indicate prediction performance on held-out cell lines and drugs. Higher AUROC and R\textsuperscript{2} denote better performance. 
Column bests: \textbf{bold} (1st), \underline{underlined} (2nd), \textdagger{} (3rd).}
\label{tab:test2}
\begin{tabular*}{\textwidth}{@{\extracolsep\fill}lcccc}
\toprule%
& \multicolumn{2}{@{}c@{}}{Classification (AUROC)} & \multicolumn{2}{@{}c@{}}{Regression (R\textsuperscript{2})} \\\cmidrule{2-3}\cmidrule{4-5}%
Method & Cell Line & Drug & Cell Line & Drug \\
\midrule
DeepDSC  & 0.693 (±0.296) & 0.668 (±0.037) & \underline{0.519} (±0.297) & \underline{-0.095} (±0.298) \\
GraphCDR & 0.346 (±0.294) & 0.588 (±0.289) & 0.162 (±0.195) & -3.902 (±7.822) \\
SubCDR   & 0.653 (±0.264) & 0.728 (±0.082) & -0.162 (±0.348) & -1.854 (±4.962) \\
DRPreter & 0.516 (±0.277) & 0.697 (±0.153) & 0.320 (±0.321) & -7.497 (±18.837) \\
MOFGCN   & \underline{0.711} (±0.262) & 0.830\textsuperscript{\textdagger} (±0.149) & 0.221\textsuperscript{\textdagger} (±0.027) & -2.300 (±57.272) \\
NIHGCN   & \textbf{0.982} (±0.025) & \textbf{0.846} (±0.181) & 0.220 (±0.033) & -0.962 (±20.553) \\
drGT     & 0.706\textsuperscript{\textdagger} (±0.198) & \underline{0.844} (±0.139) & \textbf{0.692} (±0.178) & \textbf{0.022} (±0.052) \\
\botrule
\end{tabular*}
\end{table}

Table~\ref{tab:test2} presents the average AUROC and $R^2$ scores for all models on the GDSC2 dataset under the leave-one-out evaluation setting for drugs or cell lines where we masked (i) all drug response values for a given drug across all cell lines or (ii) cell line response across all drugs. All $R^2$ values are computed using the standard coefficient of determination without any rescaling and therefore lie in the range $(-\infty, 1]$.

\textbf{Classification:}  drGT ranks among the top three, especially for unseen drugs. In the drug-clear-out setting, it achieves an AUROC within 0.002 of NIHGCN (the best-performing baseline) and outperforms MOFGCN (the second-best baseline) by 0.014 AUROC. In the cell-clear-out setting, drGT achieves an AUROC only 0.005 lower than the second-best baseline, MOFGCN. While NIHGCN remains the top performer with a substantial margin (+0.276 in AUROC over drGT), our method still ranks among the top three, demonstrating competitive performance. 

\textbf{Regression:} Under the more challenging regression setting, drGT achieves the highest average $R^2$ in both cell- and drug-level evaluations, improving on DeepDSC by +0.17 for cell-level prediction and +0.12 for drug-level prediction. In particular, the GDSC2 regression setting involves only 95 drugs but over 600 cell lines, resulting in highly heterogeneous response profiles per drug. We note that the absolute $R^2$ values for drug-level prediction remain low and often negative across all methods, which is expected under the leave-drug-out setting due to severe distribution shifts and low variance within individual held-out drugs. In this context, the transition from negative to slightly positive $R^2$ indicates improved robustness rather than strong absolute predictive accuracy.

We hypothesize that the lower performance for unseen cell lines arises from data sparsity. This effect is more pronounced in regression than in classification, where response binarization and rank-based metrics such as AUROC reduce sensitivity to response heterogeneity. For instance, the GDSC2 classification dataset contains 10,113 drug-cell response entries, with an average of 66.1 nonzero entries per drug and 15.5 per cell line, indicating fewer data points per cell line. This imbalance poses greater challenges for generalizing to unseen cell lines, as models must extrapolate from substantially fewer observations per cell line. Nevertheless, drGT maintains relatively strong performance under these conditions, likely benefiting from its kernel-based similarity representations, which mitigate the effects of data sparsity.

\subsubsection{Performance of Zero-Shot Capability (Test 3)}

\begin{table}[htbp]
\centering
\caption{
Test 3: Zero-shot evaluation on GDSC2 using models trained on GDSC1 for classification.
``Seen pairs'' are drug-cell combinations included in training, 
whereas ``Unseen pairs'' are novel combinations absent from training. 
Results are reported as AUROC with 95\% CI. 
}
\vspace{-10pt}
\label{tab:zeroshot_combined}

\begin{tabular}{c|cccc}
\toprule
Method &
\begin{tabular}[c]{@{}c@{}}Seen pairs\\(n=986)\end{tabular} &
\begin{tabular}[c]{@{}c@{}}Unseen pairs\\(n=4,219)\end{tabular} &
\begin{tabular}[c]{@{}c@{}}Overall\\(n=5,205)\end{tabular} &
95\% CI \\
\midrule
DeepDSC & 0.551 & 0.518 & 0.540 & 0.524-0.556 \\
GraphCDR & 0.754 & 0.606 & 0.643 & 0.628-0.658 \\
SubCDR & 0.726 & 0.583 & 0.630 & 0.615-0.645 \\
DRPreter & 0.730 & \underline{0.718} & 0.728 & 0.714-0.741 \\
MOFGCN & \underline{0.892} & 0.705\textsuperscript{\textdagger} & 0.766\textsuperscript{\textdagger} & 0.752-0.778\textsuperscript{\textdagger} \\
NIHGCN & \textbf{0.893} & 0.704 & \underline{0.772} & \underline{0.759-0.784} \\
drGT & 0.884\textsuperscript{\textdagger} & \textbf{0.733} & \textbf{0.786} & \textbf{0.774-0.798} \\
\bottomrule
\end{tabular}%
\end{table}

We further evaluated the model’s zero-shot capability by training it on GDSC1 (assay: Resazurin or Syto60) and testing it on GDSC2 (assay: CellTiter-Glo), which involves a different assay protocol and thus represents a distribution shift. The two datasets share 644 drugs and 83 genes, resulting in 5,205 drug-cell line response pairs, of which 986 are observed in GDSC1 and 4,219 are unseen. Models were trained on GDSC1 and evaluated on GDSC2. The AUROC with 95\% confidence intervals (CIs) were estimated using the nonparametric DeLong method~\cite{delong1988comparing}, which computes AUROC variance without assuming any specific distribution. For regression, 95\% CIs were estimated via nonparametric bootstrapping (1,000 resamples) by resampling drug-cell pairs with replacement and recalculating $R^2$ values for each replicate.
			
\textbf{Classification} Table~\ref{tab:zeroshot_combined} presents the AUROC and corresponding CIs for Test~3. Our model achieves the highest AUROC for overall, seen, and unseen pairs, followed by NIHGCN and MOFGCN. Interestingly, drGT ranks third on seen pairs but first on unseen pairs, whereas NIHGCN performs best on seen pairs but drops to fifth on unseen pairs. This suggests that our model generalizes better to out-of-distribution data, while NIHGCN may overfit to the training distribution.

\begin{figure*}[!htbp]
\centering
  \includegraphics[width=\textwidth]{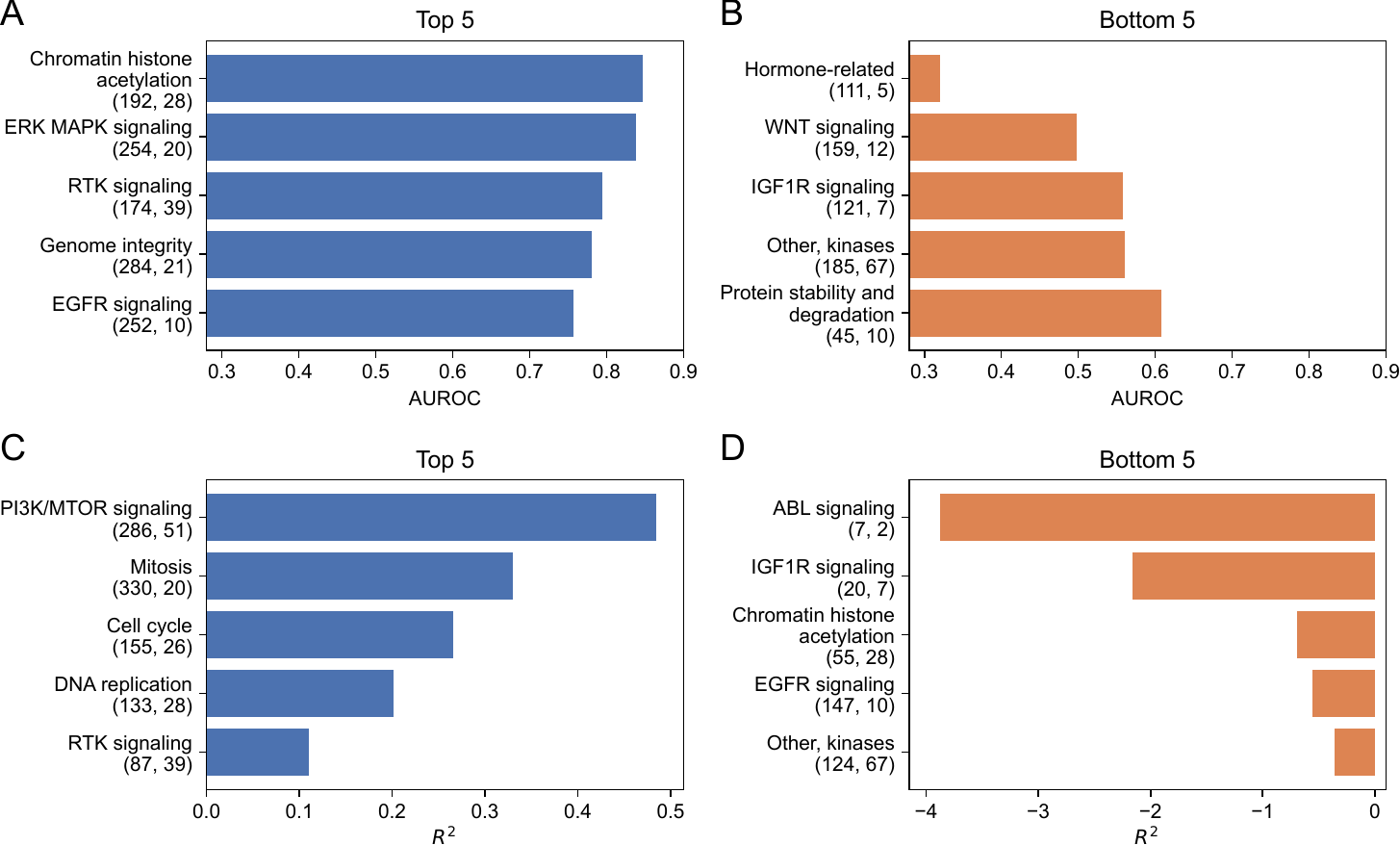}%
\caption{
Pathway-level generalization performance of drGT for unseen drugs in GDSC2.
Each panel displays the top five and bottom five target pathways, ranked by model performance. Panels (A, B) show AUROC values representing classification accuracy for unseen compounds, whereas panels (C, D) present $R^2$ values reflecting regression performance for drug response prediction. Numbers in parentheses indicate the number of unseen drug-cell line pairs and the number of unique drugs per mechanism pathway.
}
\label{fig:pathway}
\end{figure*}

Figure~\ref{fig:pathway}A and B illustrate the pathway-level AUROC of drGT on unseen drugs in the GDSC2 dataset. Panel A shows the top five pathways with the highest AUROC values, while panel B presents the bottom five. Overall, pathways in the top-performing group contain more samples and unique drugs than those in the bottom-performing group, suggesting that greater drug diversity within a pathway facilitates more robust prediction performance.

\begin{table}[htbp]
\centering
\caption{
Test 3: Zero-shot evaluation on GDSC2 using models trained on GDSC1 for regression.
``Seen pairs'' are drug-cell combinations included in training, 
whereas ``Unseen pairs'' are novel combinations absent from training. 
Values represent mean $R^2$ with 95\% CI. 
}
\label{tab:seen_unseen_overall_ci}
\begin{tabular}{c|cccc}
\toprule
{Method} &
\begin{tabular}[c]{@{}c@{}}{Seen pairs}\\(n=986)\end{tabular} &
\begin{tabular}[c]{@{}c@{}}{Unseen pairs}\\(n=4,219)\end{tabular} &
\begin{tabular}[c]{@{}c@{}}{Overall}\\(n=5,205)\end{tabular} &
{95\% CI} \\
\midrule
DeepDSC & \underline{0.329} & \underline{0.241} & \underline{0.310} & \underline{0.264-0.355} \\
GraphCDR & -0.055 & -0.000 & -0.017 & -0.023-{-0.012} \\
SubCDR & 0.269\textsuperscript{\textdagger} & 0.171\textsuperscript{\textdagger} & 0.247\textsuperscript{\textdagger} & 0.190-0.271\textsuperscript{\textdagger} \\
DRPreter & 0.243 & 0.145 & 0.222 & 0.144-0.287 \\
MOFGCN & 0.152 & 0.112 & 0.152 & 0.143-0.160 \\
NIHGCN & 0.158 & 0.138 & 0.165 & 0.065-0.094 \\
drGT & \textbf{0.354} & \textbf{0.261} & \textbf{0.334} & \textbf{0.286-0.373} \\
\bottomrule
\end{tabular}%
\end{table}

\textbf{Regression:}  Table~\ref{tab:seen_unseen_overall_ci} summarizes the zero-shot evaluation results for the regression task.
Our model consistently outperforms all baselines on both seen and unseen pairs, indicating robust generalization to out-of-distribution data. Figure~\ref{fig:pathway}C and D further show the pathway-level $R^2$ of {drGT} on unseen drugs in the GDSC2 dataset. Pathways with larger sample sizes and greater chemical diversity, such as PI3K/MTOR and RTK signaling, tend to yield higher $R^2$ values. In contrast, pathways represented by only a few drugs (e.g., ABL signaling) exhibit poor regression performance, likely due to limited training diversity and the narrow range of dose-response profiles available for these mechanisms. 

These findings highlight two complementary aspects of {drGT}. First, its graph-based representation enables effective knowledge transfer across pharmacologically related drugs, allowing performance gains for drugs targeting data-rich pathways. Second, the drop in performance for rare or sparsely sampled pathways underscores a limitation common to data-driven models — the need for sufficient variability to learn transferable molecular patterns. Together, these observations suggest that data abundance and mechanistic diversity are key determinants of regression accuracy in zero-shot settings.



\subsection{Interpretation}

\subsubsection{Evaluation of Drug-Gene Associations in Scientific Literature}

For the interpretability analysis, we focused on the NCI60 dataset, which includes annotated MoA for drugs. To ensure reproducibility and minimize variance due to stochastic training effects, we used the AC matrix from a single fold of Test 1. This yielded two AC matrices—one from the training set and one from the validation set. For downstream analysis, we used the validation AC matrix, as it is unaffected by dropout and thus provides more stable attention patterns. 

\begin{figure*}[!ht]
\centering\includegraphics[width=\textwidth,keepaspectratio]{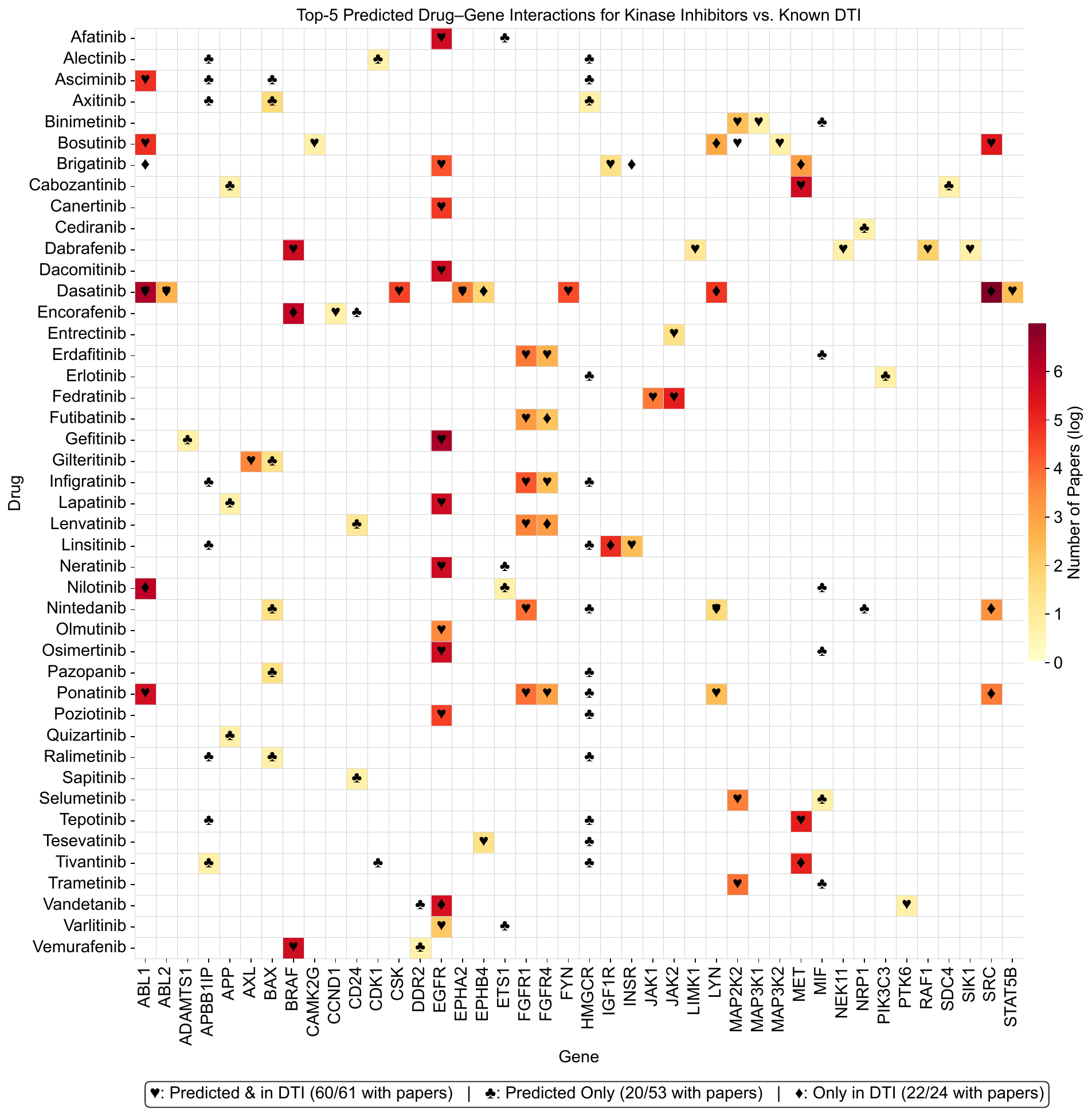}
\caption{Top-5 predicted drug-gene interactions for kinase inhibitor drugs (-nib compounds) in the NCI60 dataset, compared with known drug-target interactions (DTI) using co-occurrences based on PubMed abstracts. Drugs are ordered alphabetically, and this figure shows a subset of the data for clarity. The color represents the number of abstracts associated with a specific drug and gene pair by natural log scale. Symbols indicate the following: $\varheartsuit$ represents relationships predicted by drGT and present in the DTI dataset ($61$ instances); $\clubsuit$ represents relationships predicted by drGT but not present in the DTI dataset (53 instances); $\vardiamondsuit$ represents relationships only present in the DTI dataset (24 instances). Several drugs have 5 or more co-occurrences due to multiple NSCs. This figure shows a subset of the data for clarity.}
\label{fig:heatmap}
\end{figure*}
Figure~\ref{fig:heatmap} displays a heatmap with the drug-gene co-occurrences based on abstract text. Out of 4,880 drug-gene attention-based relationships, 356 of the drug-gene co-occurrences were found in the abstracts from 16,831 articles. We present this as a subset heatmap in Figure~\ref{fig:heatmap}, which includes only drugs mentioned by name in at least one abstract, explicitly focusing on compounds whose names end with “-nib”. For both heatmaps, the color corresponds to the number of publications identified, using a natural log scale. Likewise, the symbols in Figure~\ref{fig:heatmap} indicate different relationship types. The $\varheartsuit$ symbol represents relationships predicted by drGT and present in the DTI dataset (61 instances in the subset). The $\clubsuit$ symbol indicates relationships predicted by drGT but not present in the DTI dataset (53 cases in the subset). The $\vardiamondsuit$ symbol denotes relationships only present in the DTI dataset (24 instances in the subset). Table~\ref{tab:attention} summarizes the full attention-derived drug-gene associations supported by the literature. Across 976 drugs, we extracted 4,880 associations corresponding to the top 5 genes per drug based on ACs. Among these, 4,669 were novel (i.e., not present in the known DTI dataset, while 211 exist in the known DTI dataset (Table~\ref{tab:attention}); this is 36.9\% DTIs of the 572 DTIs included.  Of all associations, 356 had at least one supporting PubMed abstract. Among the supported associations, 182 were novel, indicating that drGT can identify literature-supported targets that are not present in current DTI datasets (we term these ``novel pairs''). Additionally, 12.2\% of the ``novel pairs''  had at least one PubMed-supported gene among their top-ranked predictions, highlighting the potential of our approach to identify targets for understudied compounds. These results demonstrate that ACs can recover known drug-gene associations and identify under-characterized pairs supported by the literature, suggesting their potential value for downstream experimental validation.

\begin{table}[t]
  \centering
  \caption{Validation of Top-5 attention-derived drug-gene associations
  using PubMed co-occurrence evidence}
  \label{tab:attention}
  \begin{tabular}{@{}l r@{}}
    \toprule
    Metric & Value \\
    \midrule
    Total Predicted Drug-Gene Pairs & 4,880 \\
    \quad Known (in DTI) & 211 \\
    \quad Novel (not in DTI) & 4,669 \\
    \midrule
    Total Predicted Pairs with PubMed Co-Occurrences & 356 \\
    \quad Novel Pairs with PubMed Support & 182 \\
    \quad Known Pairs with PubMed Support & 174 \\
    \midrule
    \shortstack[l]{Drugs with $\ge 1$ PubMed-Supported\\Top-5 Novel Pair (not in DTI)} & 12.2\% \\
    \bottomrule
  \end{tabular}
\end{table}

\subsubsection{Assessing Drug-Target Interactions from Attention Coefficients}

\begin{figure*}[!ht]
\centering
  \includegraphics[width=\textwidth]{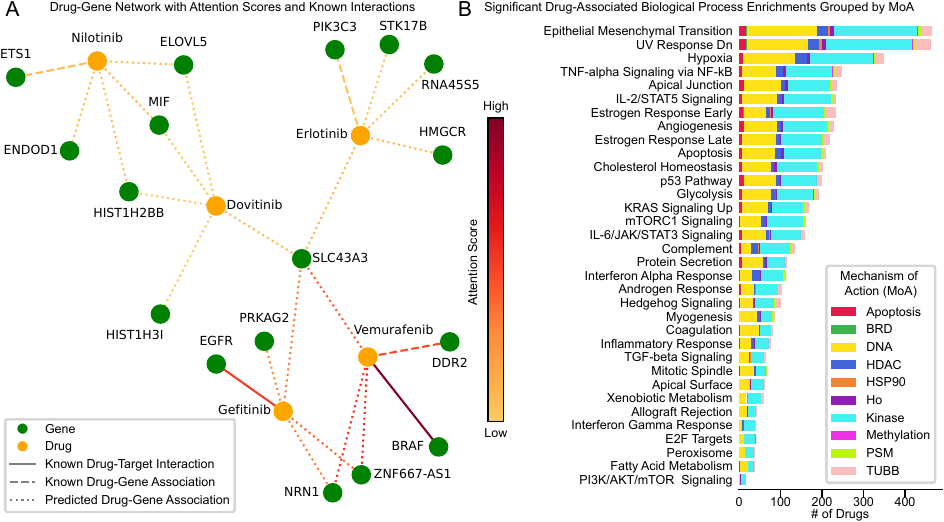}%
 \caption{Drug-gene association assessment. (A) DGA network based on ACs. The network shows relationships between drugs (orange nodes) and genes (green nodes) based on the top 5 ACs from drGT. Edges are colored according to the AC value. Lines: known DTIs (i.e., input training data; solid); known DGA (relationships described in abstracts, but not included as model input, dashed); predicted DGAs (dotted). (B) Drug-associated biological processes from gene ACs. A bar chart representing the number of drugs linked to various biological processes, determined by over-representation analysis of hallmark gene sets. X-axis: represents biological processes. Y-axis: number of drugs. The color coding of these bars corresponds to the MoA, including categories such as DNA-targeting agents, tubulin inhibitors (TUBB), kinase inhibitors, hormone-related drugs (Ho), apoptosis inducers, methylation modulators, HDAC inhibitors, HSP90 inhibitors, proteasome inhibitors (PSM), and bromodomain inhibitors (BRD)}
\label{fig:DTI}
\end{figure*}

The intricate network of DTIs and broader drug-gene associations (DGAs), as well as, the quantification of their relationships from the model results via ACs aid our further understanding of MoAs. Figure~\ref{fig:DTI}A presents a visualization of these DGAs, employing the drGT model to highlight the interplay between drugs and the genes they associate with, focusing on five kinase inhibitors: dovitinib (NSC-759661), erlotinib (718781), gefitinib (759856), nilotinib (747599), and vemurafenib (761431).

Of the kinase inhibitors shown in Figure~\ref{fig:DTI}A, gefitinib and vemurafenib are shown with links to their well-known protein targets (EGFR with L858R and BRAF with V600E mutations, respectively)~\cite{tracy2004gefitinib,bollag2012vemurafenib}. Within Figure~\ref{fig:DTI}, of the drugs with known or plausible drug-gene associations (based on co-occurrences in PubMed abstract text), we observed vemurafenib, erlotinib, and nilotinib. For vemurafenib, our model identified DDR2 (a receptor tyrosine kinase) as a predicted target. In direct validation of our observation that vemurafenib inhibits DDR2, vemurafenib has been shown to reduce DDR2 activity to less than 30\%~\cite{horbert2015photoactivatable}. With respect to erlotinib, prior research suggests that inhibition of PIK3C3 (also known as VPS34) can sensitize specific breast cancer cell lines to erlotinib, implying that the model may be capturing biologically relevant downstream or compensatory processes~\cite{dyczynski2018targeting}. Similarly, for nilotinib, which targets ABL1 and KIT~\cite{blay2011nilotinib}, the model identified ETS1, a gene associated with the BCR-ABL signaling pathway~\cite{wang2025ets}. This relates to reports that nilotinib suppresses ETS1 expression~\cite{tripathi2018abl}, suggesting that the model may capture signaling effects that contribute to the drug’s effect.

The third category of observations from the drGT predictions are recurrent genes that have received less direct investigation with individual drugs but may be of interest for future exploration. Two such genes are NRN1 (identified for gefitinib and vemurafenib) and SLC43A3 (identified for dovitinib, erlotinib, gefitinib, and vemurafenib). NRN1 is a neurotrophic protein (a subset of growth factors) that has been shown to have both tumor suppressive and oncogenic activities related to kinases~\cite{du2021methylation}. NRN1 suppresses esophageal cancer growth by inhibiting PI3K-Akt-mTOR signaling~\cite{du2021methylation}, while NRN1 was shown to have a role in oncogenic STAT3 signaling in melanoma~\cite{devitt2024nrn1}. Recent work observed worse prognoses in patients with high SLC43A3 expression. These patients with high SLC43A3 expression also showed enrichment of the PI3K‐AKT signaling pathway and enhanced EGF signaling, which may relate to erlotinib and gefitinib activity (both EGFR inhibitors)~\cite{wu2024multi}. Dovitinib, a known FGFR family inhibitor~\cite{andre2013targeting}, was associated with five genes: SLC43A3, MIF, ELOVL5, HIST1H2BB, and HIST1H3I. While these associations have not yet been experimentally explored, they may represent off-target effects or alternative pathways involved in the drug’s MoA, offering novel hypotheses for further investigation.

In summary, the model identifies a set of underexplored genes that may contribute to therapeutic response or resistance across multiple kinase inhibitors. The recurrence of specific genes, such as NRN1, ZNF667-AS1, and SLC43A3, may suggest underlying commonalities. 

While not a comprehensive analysis, the above results demonstrate that our model maintains high accuracy in classifying known DTIs used as training data and also predicts several known drug-gene relationships reported in the literature. This suggests that our predicted DGAs can indicate potential targets or drug-gene relationships suitable for study. 

\subsubsection{Over-Representation Analysis with Attention Coefficients}

We elucidate the functional roles of genes associated with various drugs by leveraging ACs derived from our model. We highlight the biological processes with p-values adjusted by Benjamini-Hochberg and show those $<$ 0.05 in Figure~\ref{fig:DTI}B, showing the drug counts associated with each process. Figure~\ref{fig:DTI}B presents the number of drugs related to each enriched biological process, grouped by their corresponding MoA categories. Among the 976 drugs analyzed, 768 exhibited enrichment in at least one biological process (FDR $<$ 0.05). These 768 drugs were categorized as follows: kinase inhibitors ($n = 322$), DNA-targeting agents ($n = 258$), HDAC inhibitors ($n = 44$), tubulin-targeting agents ($n = 40$), apoptosis inducers ($n = 25$), hormone-related agents ($n = 13$), HSP90 inhibitors ($n = 12$), proteasome inhibitors ($n = 11$), methylation modulators ($n = 4$), bromodomain inhibitors ($n = 3$), and agents with other or undefined mechanisms ($n = 36$). 

Among these categories, kinase inhibitors showed enrichment for a broad spectrum of biological processes, including epithelial-mesenchymal transition (EMT), KRAS signaling, and IL-2/STAT5 signaling, consistent with their known related pathway~\citep{cohen2021kinase, du2016targeting, beadling1996interleukin}. DNA damage drugs were enriched in the UV response down, p53 signaling, apoptosis pathways, and xenobiotic metabolism pathways. These findings align with studies linking DNA damage to transcriptional changes (UV response), p53-mediated cell cycle arrest, and induction of apoptosis~\citep{kaya2025uv, abuetabh2022dna, hientz2016role}. Similarly, the involvement of xenobiotic metabolism pathways highlights how these drugs are processed. Other MoAs showed expected associations: HDAC inhibitors with apoptosis~\cite{seto2014erasers}, hormone-related drugs with apoptosis~\cite{lewis2009estrogen}, and tubulin-targeting agents with EMT~\cite{whipple2010epithelial} and UV inhibition of microtubules assembly~\cite{zaremba1984effects}. Together, these observations highlight how ACs can capture biologically meaningful and MoA-consistent transcriptional programs.

\subsection{AC-Derived Targets Align with External Predictors Without Explicit DTI Supervision}

\begin{figure*}[!ht]
\centering
  \includegraphics[width=\textwidth]{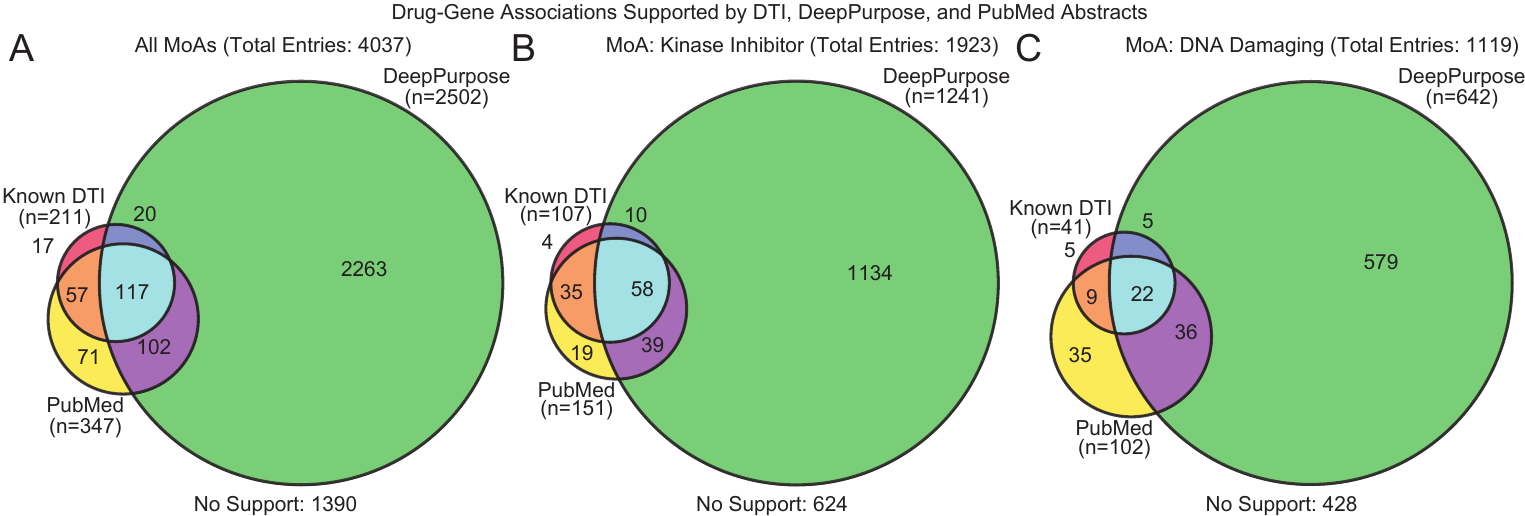}%
  \caption{
  The overlap of DGAs predicted by DeepPurpose, supported by PubMed literature, and known DTI databases (DrugBank). (A) All predicted DGAs ($n = 4{,}037$). (B) Subset of DGAs targeting kinases ($n = 1{,}923$). (C) Subset of DGAs targeting DNA ($n = 1{,}119$). Each circle represents support from one of three sources: known DTIs, DeepPurpose predictions, or PubMed evidence. The number of DGAs supported by each combination is indicated. Entries without support from any of the three sources are listed as ``No Support''.
  }
  \label{fig:compWithDP}
\end{figure*}

Figure~\ref{fig:compWithDP} summarizes how AC-derived drug-gene associations (DGAs) align with three independent sources of evidence: known DTIs from DrugBank, PubMed co-occurrence, and predictions from DeepPurpose. Of the initial 4,880 AC-derived drug-gene pairs, 4,037 were eligible for comparison after filtering out pairs lacking the information required for DeepPurpose evaluation (e.g., SMILES structure and protein sequence).

Among these 4,037 DGAs, 2,647 (65.57\%) were supported by at least one external source (Fig.~\ref{fig:compWithDP}A). A substantial fraction of associations not found in DrugBank nevertheless aligned with one or both alternative evidence sources: 2,436 of 3,826 novel DGAs (63.67\%) were supported by either PubMed co-occurrence or DeepPurpose predictions. Notably, 102 novel DGAs were supported by both resources, representing high-confidence candidates for future validation, while 71 were supported by PubMed alone. Conversely, 2,263 associations were supported only by DeepPurpose, suggesting they may represent computationally inferred but as-yet unreported associations.

We next examined evidence support rates across mechanisms of action (MoAs) with many entries. Kinase inhibitors exhibited the highest support, with 65.64\% (1,192/1,816) receiving support from at least one evidence source (Fig.~\ref{fig:compWithDP}B). DNA-targeting agents showed slightly lower agreement (60.3\%, 650/1,078; Fig.~\ref{fig:compWithDP}C), consistent with their more heterogeneous and less well-characterized interaction profiles.

Finally, 34.43\% of all AC-derived DGAs had no support from any external resource. These unsupported associations may reflect false positives or previously uncharacterized drug-gene relationships that warrant further experimental investigation.

\section{Discussion}\label{sec12}
We developed the drGT model to derive attention coefficients (ACs) for each node, including genes, drugs, and cell lines. This approach can predict both binary responses and negative log IC50 () values, as well as, facilitate model interpretation. Furthermore, our model achieves competitive or SOTA performance in randomly masked, unseen drug/cell settings and zero-shot learning. Our model provides new connections between drugs and genes beyond the original DTI training examples. For non-DTI drug-gene relationships, our analysis identified potentially novel associations using our AC-based interpretability approach, not found in other SOTA models. This capability enables our model to produce accurate response predictions while identifying novel drug-gene relationships that may help in understanding drug mechanisms.

A limitation of this study, as with other benchmarked methods, is the lack of a clear mechanistic description of how genes influence drug response. This level of uncertainty also exists in many biological studies that provide data resources that compress dimensions of magnitude, direction, and mechanism in how entities interact, but this type of compression is common. Databases, such as the Comparative Toxicogenomics Database, include interactions in their data collections, as they can serve as the basis for more detailed studies~\cite{davis2025comparative}. Another challenge in this study is the amount of high-confidence DTI data available. For example, although the NCI60 dataset includes 976 compounds, only 191 drug compounds have DrugBank DTI data. This highlights the need for further collection of drug-gene relationships available for this type of study. Ongoing efforts aim to expand such datasets. This expansion leverages advancements in information extraction and natural language processing, as well as crowd-sourced curation~\citep{wong2021science, rodchenkov2020pathway}. 

\section{Conclusion}

In conclusion, this work highlights how results from ML techniques can be paired with bioinformatic analyses to understand mechanisms of cellular response to treatment. We expect future work to broaden the approach while providing more granularity to its interpretability. Currently, our model leverages drug response data to compute cell line similarity. Future work could incorporate additional multi-omics data, including methylation, copy number variation, and mutation data, as has been done by others~\citep{peng2021predicting}. We believe this enhancement will both boost model performance and interpretability. 

\backmatter
\section*{Declarations}

\subsection*{Ethics approval and consent to participate}
Not applicable. This study did not involve human participants, human data, or animal experiments.

\subsection*{Consent for publication}
Not applicable. This study does not include any person’s data.

\subsection*{Data Availability}
The datasets supporting the conclusions of this article originated from the following public sources: the NCI60 dataset is available in the CellMinerCDB repository (\url{https://discover.nci.nih.gov/cellminercdb/}); the GDSC dataset is available in the CancerRxGene repository (\url{https://www.cancerrxgene.org/}); the CTRP dataset is available in the Broad Institute repository (\url{https://portals.broadinstitute.org/ctrp/}); the PharmacoDB dataset is available at (\url{https://pharmacodb.ca/}), and the DrugBank dataset is available in the DrugBank repository (\url{https://go.drugbank.com/}). Our processed datasets used for training are available at: \url{https://zenodo.org/records/18685976}

The code supporting the findings of this study (drGT) is available at \url{https://github.com/sciluna/drGT}.

\subsection*{Competing interests}
The authors declare that they have no competing interests.

\subsection*{Funding}
The authors received funding from the Google Summer of Code program and support from the Division of Intramural Research (DIR) of the National Library of Medicine (NLM), National Institutes of Health (NIH) (ZIALM240126). The funders were not involved in any aspect of this research.

\subsection*{Authors' contributions}
Y.I. conceived and designed the study, implemented the methods, performed the experiments, and analyzed the results. H.L. contributed to the implementation and experimental evaluation. A.L. and R.K. contributed to the experimental design, evaluation, and discussion of the results, and provided feedback on manuscript writing. T.F. assisted with proofreading and manuscript editing. All authors read and approved the final manuscript.

\subsection*{Acknowledgements}
We would like to acknowledge helpful conversations with Vinodh N. Rajapakse.

\begin{appendices}

\section{drGT Flow Chart} \label{app:flow_chart}

\begin{figure*}[!ht]
\centering
\includegraphics[width=\textwidth,height=0.8\textheight,keepaspectratio]{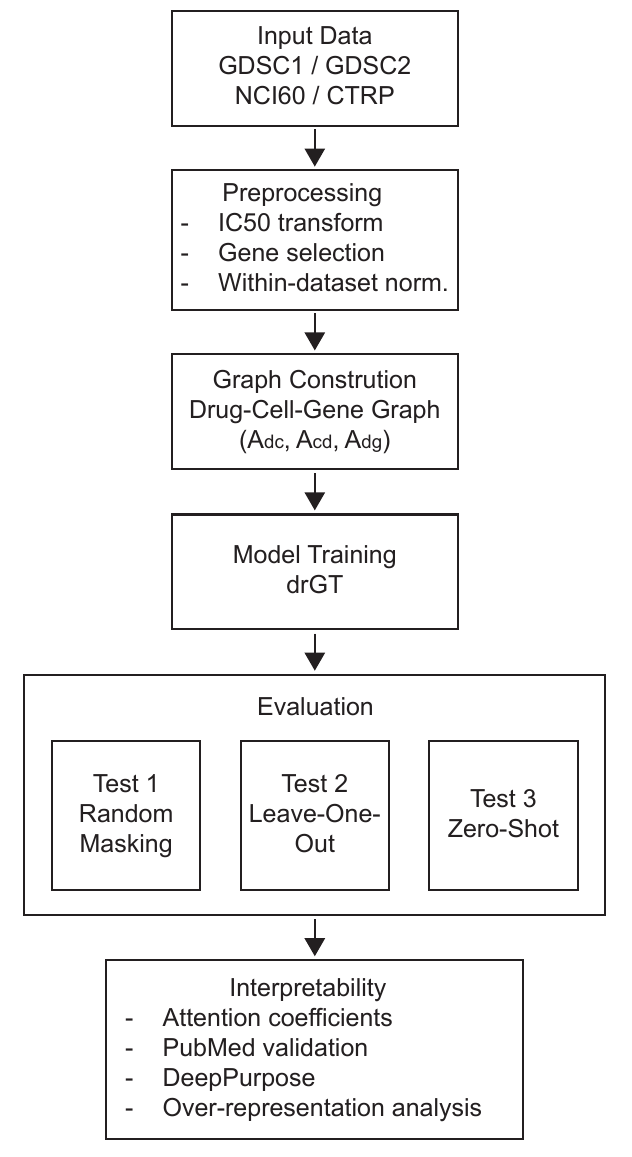}
\caption{%
Overview of the experimental workflow.
The flowchart summarizes data preprocessing, heterogeneous drug-cell-gene graph construction, drGT model training, and evaluation under three experimental settings (random masking, leave-one-out, and zero-shot transfer), and downstream attention-based interpretability analyses.}
\label{fig:flow_chart}
\end{figure*}

Figure~\ref{fig:flow_chart} provides a high-level overview of the experimental workflow used in this study. Starting from multiple pharmacogenomic datasets (i.e., GDSC1, GDSC2, NCI60, and CTRP) the raw data are first preprocessed by applying dataset-specific normalization, IC50 transformations, and gene selection procedures.

Based on the preprocessed data, a heterogeneous drug-cell-gene graph is constructed to encode drug response relationships as well as biological side information. The drGT model is then trained on this graph using observed drug-cell associations. Model performance is evaluated under three different experimental settings: random masking of drug-cell pairs, leave-one-out evaluation for drugs or cell lines, and zero-shot transfer across datasets.

Finally, attention coefficients learned by the Graph Transformer layers are used for downstream interpretability analyses, including literature-based validation with PubMed drug-gene co-occurrence, independent computational validation using DeepPurpose, and over-representation analysis to assess biological plausibility.

\section{Hyperparameter Search Space} \label{app:hyperparam}

This section describes the hyperparameter search space explored using Optuna during the model development process. 

\begin{itemize}
  \item \textbf{GNN layer dimensions:} $(h_1, h_2, h_3)$ where $h_1 \in [256, 512]$, $h_2 \in [64, \min(256, h_1)]$, $h_3 \in [32, \min(128, h_2)]$
  \item \textbf{Number of GNN layers:} $\{2, 3, 4\}$
  \item \textbf{Attention heads:} $\{2, 3, \dots, 8\}$
  \item \textbf{Dropout (pre/post/MLP):} 0.1-0.5
  \item \textbf{MLP layers:} 1 to 3
  \item \textbf{Activation function:} ReLU or GELU
  \item \textbf{Optimizer:} Adam or AdamW
  \item \textbf{Normalization:} GraphNorm, BatchNorm, or LayerNorm
  \item \textbf{Learning rate:} log-uniform in $[10^{-5}, 10^{-2}]$
  \item \textbf{Weight decay:} log-uniform in $[10^{-6}, 10^{-2}]$
  \item \textbf{Epochs:} $\{300, 400, \dots, 1500\}$
  \item \textbf{Scheduler:} Optional cosine annealing (with $T_\mathrm{max}$ set to total epochs)
  \item \textbf{Attention dropout:} 0.0-0.4
\end{itemize}

The hyperparameter optimization aimed to minimize validation loss while maintaining stable convergence across datasets and random seeds.

\section{Comparison between Zero Padding and Non-Zero Padding}

We also evaluated the impact of zero padding on the representation of drug-gene interactions. Zero padding was applied to standardize input feature dimensions across samples in heterogeneous datasets. Table~\ref{tab:transformer_zero_pad} summarizes the results, showing that models with zero padding generally achieved higher AUROC values, particularly on the NCI60 and CTRP datasets.

\begin{table}[t]
  \centering
  \caption{Performance comparison of Transformer models with and without zero padding on each dataset.}
  \label{tab:transformer_zero_pad}
  \begin{tabular}{@{}lcc@{}}
    \toprule
    \textbf{Dataset} & \textbf{Without Zero Padding} & \textbf{With Zero Padding} \\
    \midrule
    CTRP   & 0.897 & 0.935 \\
    GDSC1  & 0.869 & 0.884 \\
    GDSC2  & 0.887 & 0.897 \\
    NCI60  & 0.820 & 0.946 \\
    \bottomrule
  \end{tabular}
\end{table}

\end{appendices}


\bibliography{sn-bibliography}

\end{document}